\documentclass[10pt,journal,compsoc]{IEEEtran}
\ifCLASSOPTIONcompsoc
\usepackage[nocompress]{cite}
\else
\usepackage{cite}
\fi
\ifCLASSINFOpdf
\usepackage[pdftex]{graphicx}
\graphicspath{{images/}}
\DeclareGraphicsExtensions{.pdf,.jpeg,.png}
\else
\fi
\usepackage{amsmath}
\ifCLASSOPTIONcompsoc
\usepackage[caption=false,font=footnotesize,labelfont=sf,textfont=sf]{subfig}
\else
\usepackage[caption=false,font=footnotesize]{subfig}
\fi

\usepackage{stfloats}
\usepackage[normalem]{ulem}
\usepackage{xcolor}
\usepackage{paralist}
\usepackage{scalerel}
\usepackage{hyperref}
\usepackage{bm}
\usepackage[locale=FR]{siunitx}

\newcommand{\clc}[1]{\text{#1}}

\newcommand{\glf}[1]{\text{#1}}

\usepackage{multicol}
\newcommand{\nScans}{n}

\newcommand{\nFolds}{k}
\newcommand{\nTrials}{t}
\newcommand{\nTopFeats}{m}

\newcommand{\Expert}[1]{\textit{E#1}}

\newif\ifShowEdits

\ifShowEdits
\definecolor{blue}{rgb}{0,0,1}
\renewcommand{\sout}[1]{\unskip}
\else
\definecolor{blue}{rgb}{0,0,0}
\renewcommand{\sout}[1]{\unskip}
\fi

\begin{document}
	%
	\title{A Predictive Visual Analytics System for Studying Neurodegenerative Disease Based on DTI Fiber Tracts}
	
	%
	%
	%
	%
	
	\author{
		Chaoqing~Xu,
		Tyson~Neuroth, 
		Takanori~Fujiwara, 
		Ronghua~Liang, 
		and~Kwan-Liu~Ma
		
		\IEEEcompsocitemizethanks{\IEEEcompsocthanksitem C. Xu and R. Liang are with College of Computer Science, Zhejiang University of Technology.\protect\\
			E-mail: superclearxu@gmail.com, rhliang@zjut.edu.cn.
			\IEEEcompsocthanksitem T. Neuroth, T. Fujiwara, and K.-L. Ma are with University of California, Davis.\protect\\
			E-mail: \{taneuroth, tfujiwara, klma\}@ucdavis.edu.\\
			\IEEEcompsocthanksitem Ronghua~Liang is the corresponding author\\}
		\thanks{Manuscript received September xx, 2020; revised xxx, 2020.}
	}

	%
	
	%

	\markboth{Journal of \LaTeX\ Class Files,~Vol.~14, No.~8, August~2015}%
	{Shell \MakeLowercase{\textit{et al.}}: Bare Demo of IEEEtran.cls for Computer Society Journals}
	%



	\IEEEtitleabstractindextext{%
		\begin{abstract}
			Diffusion tensor imaging (DTI) has been used to study the effects of neurodegenerative diseases on neural pathways, which may lead to more reliable and early diagnosis of these diseases as well as a better understanding of how they affect the brain. We introduce a predictive visual analytics system for studying patient groups based on their labeled DTI fiber tract data and corresponding statistics. The system's machine-learning-augmented interface guides the user through an organized and holistic analysis space, including the statistical feature space, the physical space, and the space of patients over different groups. We use a custom machine learning pipeline to help narrow down this large analysis space and then explore it pragmatically through a range of linked visualizations. We conduct several case studies using  DTI and T1-weighted images from the research database of Parkinson's Progression Markers Initiative.
		\end{abstract}
		
		\begin{IEEEkeywords}
			Brain fiber tracts, neurodegenerative disease, machine learning, predictive visual analytics, visualization.
	\end{IEEEkeywords}}

	\maketitle

	\IEEEdisplaynontitleabstractindextext

	%
	\IEEEpeerreviewmaketitle

	\IEEEraisesectionheading{\section{Introduction}\label{sec:introduction}}

	\IEEEPARstart{M}{ore} than $6.1$M people are living with Parkinson's disease (PD) ~\cite{rocca2018burden} and more than $43.8$M people are living with Alzheimer’s disease~\cite{nichols2019global}.
There is no cure~\cite{heemels2016neurodegenerative}, yet treatment can slow the progression. Thus, early detection is important but difficult due to uncertainty in how the disease begins and what the early markers are~\cite{poewe2017parkinson}; indeed, the early-stage clinical PD misdiagnosis rate is about $50\%$~\cite{mollenhauer2017depressed}.

Diffusion tensor imaging (DTI) is an advanced brain imaging technique that measures water diffusivity in brain tissue~\cite{basser2002diffusion}. By applying fiber reconstruction~\cite{SMITH20121924} to the measured diffusivity, we can estimate the location and orientation of the brain's white matter fiber tracts. 
In-depth analysis of the fiber tracts helps researchers better understand these diseases and their progression~\cite{zheng2014dti}. 
Neuroscientists have found statistical differences between healthy and diseased brains~\cite{zhang2015diffusion,acosta2016whole} using DTI images and fiber tracts. Also, many categories of machine learning (ML) techniques have been validated and employed in a range of neurodegenerative diseases~\cite{mateos2018structural,tanveer2020machine}. The use of diffusion tensor features and fiber connectivities has a potential for improving diagnostic accuracy in clinical assessment.
While such approaches can identify useful statistical markers, understanding of the relationship between the statistical and physiological features is needed to advance our knowledge of the disease.

To accomplish this, one first needs to find the relevant statistical features. 
This can be difficult since there are many irrelevant differences and datasets often lack the scale and variation to make confident statistical inferences from each feature.
In addition, these statistics are usually derived through spatial aggregation; for example, \sout{aggregating tensor measures over a canonical partition through functional region mapping.} aggregating diffusion measures over a neuroanatomical brain parcellation derived from \textcolor{blue}{anatomical landmarks} \sout{structural or functional information}. The aggregated statistics may then be used for a pairwise comparison between individuals. This is a practical approach; however, it provides a limited ability to precisely locate the physical features. For example, in a highly affected brain region, the salient statistical feature may be ``watered down'' because it may incorporate a certain amount of unaffected parts along with the affected parts. 
Visual analytics (VA) could thus add insight and confidence into the differences once the embedded physical features are discovered.

Thus, direct rendering of the fiber tracts is indispensable in providing a deeper understanding (both physiological and statistical). In addition to the fiber microstructure, color can be mapped to the fibers to display the tensor measurements (that were aggregated for statistical modeling) in their full detail. Through this process, neuroscientists may notice patterns and anomalies as well as issues that might affect the statistical analysis in non-trivial ways. They can then employ expert knowledge to reason about the statistical features and the biological factors to form a new hypothesis.

Still, after a salient statistical feature is found, its distribution in the physical space may have a high amount of additional variations between individuals. In this case, one should consider the multiple comparison problem~\cite{benjamini2010simultaneous,zgraggen2018investigating}, which highlights the risk of false visual insight discovery. Thus, the detailed physical differences must be conceptually reasoned about by experts. To maximize the effectiveness of their analysis, it is important to allow them to easily explore different features and subjects in a pragmatic way, grasp the wider context, and maintain a strong sense of awareness about the involved uncertainties. 

Based on the identified problems and through consulting neuroscientists, we design an intelligent VA system. 
To the best of our knowledge, this is the first predictive brain fiber VA system that helps neuroscientists explore neurodegenerative diseases. The user interface (UI) organizes the analysis space into three primary modalities (different modes of analysis prioritization) for exploration: statistical features, spatial regions, and individual subjects. A custom ML pipeline is used to estimate the measures of saliency and uncertainty for each modality. Specifically, we employ Extremely Randomized Trees to estimate scores for each attribute, Support-vector Machine (SVM) to predict the importance of each modality, and Cross-Validation (CV) to cope with the overfitting.
This ML-enhanced UI approach guides the user to drill down into details. Linked visualizations are used to add important context and awareness throughout the process, such as ``how good the model is'', ``how certain the model is about the suggestions'', ``how the features relate to the model'', ``how the features relate to each other'', ``how the model relates to the subjects'', ``how the groups differ with one another'', ``how individual subjects differ with others'', ``how the salient features change over time'', and ``how the features appear when spatially disaggregated and examined through direct 3D rendering''. 

Since all of the information related to the analysis is automatically displayed and linked throughout the process, one can immediately inspect and relate different aspects of the data. This provides advantages over the other systems in terms of analysis efficiency, helps to mitigate the risk of making false insights, and supports quick and informed hypothesis generation. Our specific contributions include:

\begin{compactitem}
	\item a tailored and holistic brain fiber visualization system for studying neurodegenerative disease,
	\item an ML assisted visualization pipeline to narrow down the large information space, and
	\item an exploratory analysis workflow with complimentary visualizations and interactions.
\end{compactitem}

\section{Background and Related Work}
\label{sec:backgroundAndRelatedWork}

We address three major problems in neurodegenerative disease analysis, which are identified through reviewing literature and consulting neuroscientists. We describe each problem's details, using PD research as a concrete example.
Then, we discuss relevant works.

\subsection{Problems in Neurodegenerative Disease Research}
\label{sec:problems}

We identified research challenges in the fiber-tract-based analysis of neurodegenerative disease with an extensive survey on related works. 
First, many features identified using tractography have not been thoroughly verified in comparisons with the underlying anatomy in clinic research~\cite{thomason2011diffusion}. 
Researchers also pointed out the need for a better understanding of the physiological changes in different brain regions and the detection of brain regions more focally using white matter fiber tracts~\cite{johnson2013neuromodulation, hampton2019substance}. 
Additionally, to understand the disease, it is essential to not only identify the differences between patients and controls but also examine the influential factor on disease progress from the differences within a patient group~\cite{thomason2011diffusion, hampton2019substance}. 
Lastly, there is a necessity to provide engineering guidelines to systematically explore white matter integrity in neurodegenerative disease~\cite{gold2012white}. 
Based on these identified challenges, we then summarized the following requirements.

\vspace{3pt}
\noindent\textbf{Problem 1. Large feature space.}
When analyzing neurodegenerative disease, neuroscientists typically analyze specific brain \sout{functional} regions, fiber statistics, and tensor measures. 
The whole brain can be decomposed into dozens to hundreds of regions, and the features can be extracted separately for each region~\cite{zheng2014dti}. 
The total feature space is very large when compared to the number of available calibrated brain scans~\cite{thomason2011diffusion}.
This presents statistical challenges when identifying which features are true indicators of the disease.

\vspace{3pt}
\noindent\textbf{Problem 2. Fiber microstructure and DTI measurements.}
Structural and statistical differences have been reported in many brain regions~\cite{zhang2015diffusion, aarabi2015statistical,wei2016combined,xu2021vector}. Aarabi et~al. suggested that fibers interconnecting multiple lobes may be especially atrophied and the corresponding fiber volume and average length could indicate damage~\cite{aarabi2015statistical}. 
Some studies suggested that posterior cortical atrophy begins in the left hemisphere before the right~\cite{claassen2016cortical}. 
Other researchers found that PD begins in specific regions but ultimately affects the whole brain~\cite{olde2013disrupted,yau2018network}. 
However, different studies sometimes come to contradictory conclusions~\cite{zhang2015diffusion, wen2016white}. Overall, we still do not completely understand the anatomical changes.

\vspace{3pt}
\noindent\textbf{Problem 3. Hypothesis generation.}
Neuroscientists often make hypotheses based on observations, experimental studies, and literature reviews. For example, Kamagata et~al. hypothesized that structural changes in the nigrostriatal area may indicate PD~\cite{kamagata2016neurite}.
Hepp et~al. hypothesized that damage to the fibers connecting the nucleus basalis and the cerebral cortex may cause the hallucinations suffered by PD subjects~\cite{hepp2017damaged}. 
One question that remains unsolved is the true cause of the disease.
Important factors may still be undiscovered.
Thus, new hypothesis generation is an important part of the research effort. 

\subsection{Related Work}

\textbf{Fiber tract visualization.}
MRtrix3~\cite{TOURNIER2019116137} is a state-of-the-art package for fiber reconstruction and analysis.
For visualizing fiber tracts and brain images, MITK~\cite{Fritzsche2012MITKDI} is a commonly used toolkit. SlicerDMRI~\cite{norton2017slicerdmri} is an actively maintained and widely used open-source plugin in 3DSlicer~\cite{Kikinis2014}, which is used for diffusion MRI analysis and tractography data visualization.
These tools provide a core functionality for generating and visualizing fiber tracts. 
Schultz and Vilanova provided various visualization methods, such as glyph representation of diffusion tensors and rendering of fiber tractography~\cite{https://doi.org/10.1002/nbm.3902}.
Zhang et~al. also designed glyphs for comparative visualization of the diffusion tensors~\cite{zhang2015glyph}.
However, due to the scale and complexity of fiber tract data, VA of fiber structure is still an active research topic~\cite{everts2015exploration}.

For enhanced fiber rendering, the researchers developed screen-space ambient occlusion (SSAO)~\cite{mittring2007finding} and LineAO~\cite{eichelbaum2013lineao}, which are high-performance shadow-like rendering methods and enhance spatial perception. 
Everts et~al. introduced an illustrative rendering method that emphasizes fiber structure using depth-dependent halos~\cite{everts2009depth} and a method based on a local contraction of fiber bundles to reduce occlusion while preserving the fiber macro-structure~\cite{everts2015exploration}. 
Jianu et~al. developed a method that links the 3D view with abstract 2D views. Their method helps users navigate complex fiber structure and connectivity~\cite{jianu2009exploring}. They also introduced 2D neural map projections as abstract anatomical representations~\cite{jianu2012exploring}.
Murugesan et~al.\cite{murugesan2017brain} developed a visualization tool for exploring the modular and hierarchical organization of brain regions. 

\vspace{3pt}
\noindent\textbf{Visual analytics for cohort studies.}
The broader topic that our work falls into is VA for cohort studies, where systems often require customization for specific applications. 
Preim et~al. provided a survey of this topic~\cite{doi:10.1111/cgf.13891}. Angelelli et~al. introduced a data-cube model to link heterogeneous data and help neuroscientists relate information~\cite{angelelli2014interactive}. 
Steenwijk et~al. introduced a hypothesis-driven VA framework for multi-variate, multi-model, and multi-timepoint data to facilitate across-subject visual exploration~\cite{steenwijk2010integrated}. 
Agus et~al. provided a framework that uses radiance-based absorption maps and node-link layout representations for visual exploration of energy absorption in nanometric brain volumes~\cite{https://doi.org/10.1111/cgf.13700}. 
Krueger et~al. presented a semi-automated analytics tool for interactive VA of phenotype in high-dimensional image data ~\cite{8827951}. 
For group-level feature analysis, contrastive-learning-based VA methods were developed to identify features highly contributed to each group's characteristics~\cite{8805461,zhang2021visual,fujiwara2021interactive}.  
Fujiwara et~al. developed a VA system that visualizes the similarities of multiple brain networks with dimensionality reduction~\cite{fujiwara2017visual}. 
Yang et~al. introduced a blockwise abstraction of brain connectome ensembles~\cite{yang2017blockwise}.
Angulo et~al. developed a web-based brain data visualization framework~\cite{angulo2016multi}, where they use a linked-card infrastructure for interactive filtering and view linking (similar to VTK filter pipelines~\cite{schroeder2004visualization}).
Daniel et~al. created a VA system for comparative analysis of fMRI data between subject groups~\cite{bm.20191232}. 
Daniel et~al.'s system is similar to ours in that it incorporates spatial localization, group-level comparison, linked information visualization, and a 3D anatomical view. 
However, our system (1) focuses on fiber tract data rather than fMRI data, (2) identifies regions and features with advanced ML methods, and (3) incorporates individual-level subject prediction as the third modality of exploration.
Overall, we found that VA systems tailored specifically for the detailed group-level analysis of fiber tracts are missing and there has been little VA work utilizing ML for brain data analysis.

\vspace{3pt}
\noindent\textbf{Vis+AI.} Visualization combined with ML/artificial intelligence (Vis+AI) has gained more interest recently. 
Hohman et~al. conducted an interrogative survey~\cite{8371286}.
Levy-Fix et~al. reviewed what is needed in Vis+AI to support clinical applications~\cite{levy2019machine}. 
Other examples can be found in study of semantic features in documents~\cite{ji2019visual}, high-dimensional phenotype analysis~\cite{8827951}, AI driven graph visualization~\cite{8017580,8805452}, in-situ image prediction for scientific simulations~\cite{he2019insitunet}, and automated annotation of visualizations~\cite{lai2020automatic}. 
ML also plays an important role in visualization as a basis for ranking features~\cite{mumtaz2015visualisation,maniyar2006data} or visual components~\cite{10.1145/3126594.3126653}.

\vspace{3pt}
\noindent\textbf{Predictive analysis of neurodegenerative disease.}
In neuroscience, ML has shown promise for disease detection using multiple types of data, including diffusion tensor features and anatomic fiber connectivities. 
A survey of ML applications on MRI data provides a comprehensive view of ML usages in a wide range of diseases~\cite{mateos2018structural}.
Dinov et~al. tested a variety of classification models (e.g., SVM and Decision Tree) with PD data and the results showed a significant power in predicting PD~\cite{dinov2016predictive}. Lella et~al. provided a comparison between several ML methods for neurodegenerative disease prediction using diffusion tensor measures and structural features of brain fiber tracts~\cite{10.1117/12.2274140}. 
Similarly, Castellazzi et~al. evaluated multiple ML algorithms on AD and showed great potential for improving diagnostic accuracy in clinical assessments when using the features of local diffusion tensors and \sout{functional} \textcolor{blue}{ brain region} connectivities~\cite{castellazzi2020machine}. Martin et~al. also presented that PD can be predicted with ML models using diffusion tensor measurements and white matter volumes~\cite{doi:10.1111/jon.12214}.
Despite the promise as shown in the works above, not much research has been done in the collective usage of ML and VA, which is especially intriguing for fiber tract data as it can offer unique physiological insight through qualitative VA. 
This has inspired our ML-guided VA that explores fiber tract data between subject groups. 

	\begin{figure}[t]
	\centering
	\includegraphics[width=\linewidth]{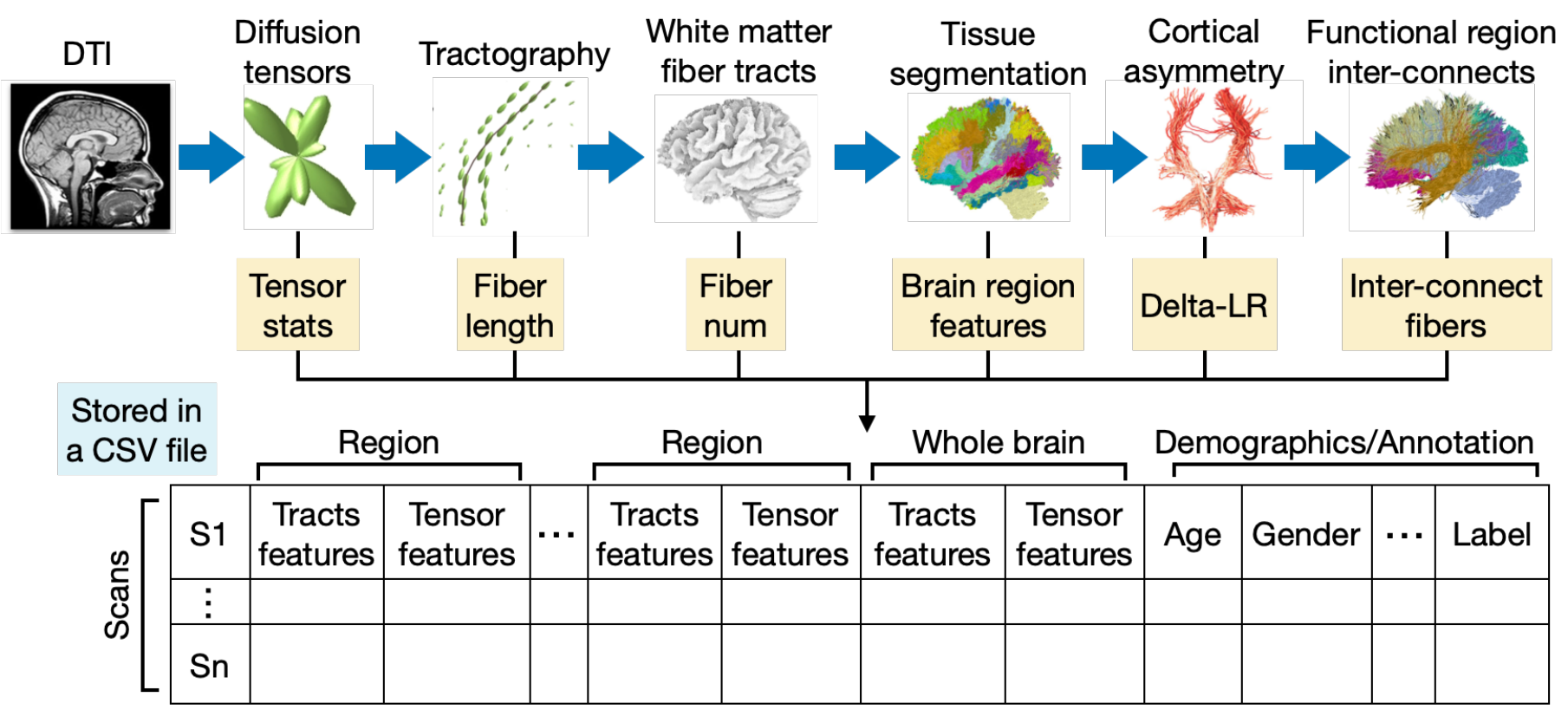}
	\caption{
		The feature extraction process and the cohort data stored in a CSV format. 
		The extracted features are described in \autoref{sec:dataDescription}.
	}
	\label{fig:Extraction}
\end{figure}

\section{Design Goals}
\label{sec:goals}

From the challenges discussed in \autoref{sec:problems}, we identify our design goals using our team's expertise, which includes ML, VA, and fiber tract visualization. 
The resulting design goals below inform our choice of methodology in \autoref{sec:methods}.

\vspace{3pt}
\noindent\textbf{DG 1: Guided analysis based on three modalities.}
The brain fiber data is multifaceted. It contains multiple features per space-time location. 
As shown in \autoref{fig:Extraction}, features extracted from the data are also high dimensional and multifaceted (e.g., features from diffusion tensors and fiber tracts).
To facilitate an effective workflow, with computational analysis support (specifically ML), our system should help the user prioritize more salient (1) features, (2) regions, and (3) subjects to choose data subsets for detailed analysis. The regions and features should be standard and interpretable so that experts can easily grasp the physiological basis, assimilate existing literature, and make hypotheses. Due to a large feature space relative to the number of scans, we must strive to avoid overfitting, reduce ranking instability, and highlight the uncertainties.

\vspace{3pt}
\noindent\textbf{DG 2: Quality visualization of brain fibers.}
For anatomical understanding, VA of the fiber tracts and salient variables in the physical space is required. 
High-quality graphics rendering can help understand the spatial relations of brain fibers. 
While the rendering should be effective at showing the structure, it should be also efficient enough to interactively render multiple large fiber sets.

\vspace{3pt}
\noindent\textbf{DG 3: Modalities for comparison.}
To address the hypothesis generation, it is fundamental to perform the comparison of different subjects and/or groups with different modalities and time steps.
Through the comparison, experts can relate diverse aspects to clinical outcomes (e.g., healthy vs PD).
As we utilize ML for the comparison, the system should also depict the relationships between data and the prediction. 

\vspace{3pt}
\noindent\textbf{DG 4: Easy non-linear exploration.}
Due to a large number of features, fiber tracts, and subjects, diverse aspects could be explored. 
The analytical process may proceed and change according to emerged patterns or discovered knowledge during the analysis. 
Therefore, our system should provide an intuitive and fully interactive UI to serve the neuroscientists' changing analysis needs.

	\begin{figure*}[ht]
	\centering
	\includegraphics[width=\linewidth]{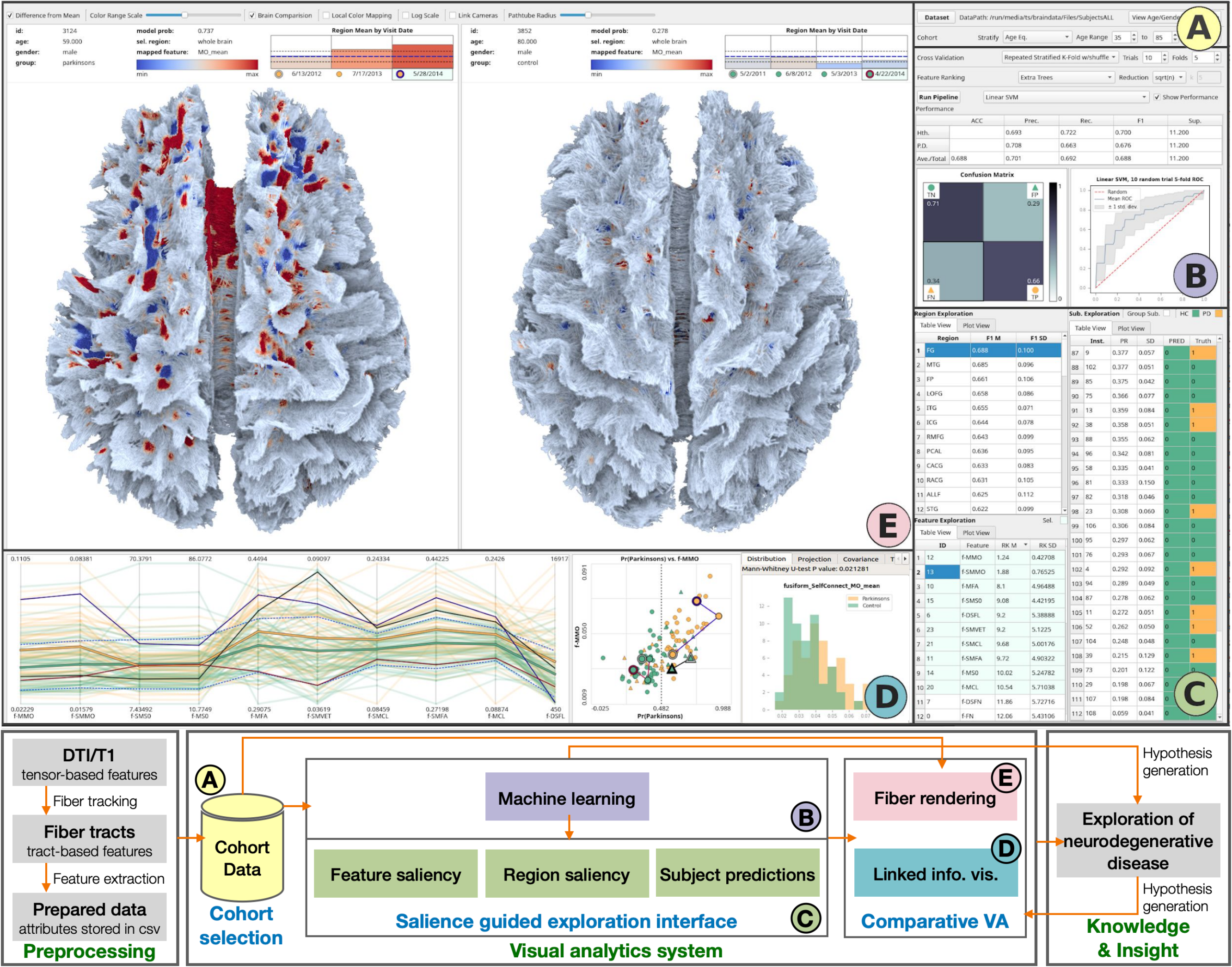}
	\caption{The UI of our VA system (top) and workflow (bottom), where each analysis step is annotated with the label of the corresponding system module. 
		The cohort selection module (\clc{A}) supports balanced and stratified subject group selection and demographic analysis. 
		The ML module (\clc{B}) can be customized before being executed. 
		After the execution, this module shows a summary of the model performance and overall uncertainty with an ROC curve, confusion matrix, the group sizes, and multiple performance measures (accuracy, precision, recall, and $F_1$). 
		The interface for exploration (\clc{C}) includes three exploration modules: the feature, region, and subject modules. 
		Each of them can be toggled between plot and table views. 
		The values displayed include the averages and standard deviations of the estimated saliencies and predictions. 
		The subject module also encodes the class labels and binary predictions. 
		The information visualization module (\clc{D}) includes a range of views for comparative analysis. 
		The 3D fiber rendering module (\clc{E}) shows selected subjects' fibers for physiological analysis.  
		The selected feature is mapped to the fibers through color. 
		Here, a divergent color shows the feature value difference between the selected subjects and the mean of the control group. 
		At the top, this module also shows the subjects' information and a timeline of their mean values of the selected feature over time. 
		The intervals on the timeline can be clicked to render the fibers corresponding to the clicked time. 
	}
	\label{fig:system}
\end{figure*}

\section{Methods} 
\label{sec:methods}

We begin with an overview of our system. {\color{blue}Then, in order to better prepare the reader to understand the system, we describe the data that we have applied our system to, including: acquisition, prepossessing, fiber tracking, and feature extraction (note that the acquisition and derivation of this data is neither novel, nor a contribution of this work, and may be changed depending on application).} Then, we describe the system's workflow, including the ML pipeline,  and interactive visualizations. 
The ML pipeline plays as the central part of the system. 
A demonstration video of the system is available at \url{https://takanori-fujiwara.github.io/s/ml-brain-fiber/}.

\subsection{System Overview}
\label{sec:systemoverview}

\autoref{fig:system} shows our VA workflow and UI. The preprocessing step of the VA workflow includes fiber tracking and feature extraction. 
Use of the system begins by selecting cohorts through a module that balances DTI scans in each group (i.e., PD or healthy control (HC)) while stratifying with age and gender of subjects corresponding to the scans (\autoref{fig:system}\clc{A}). 
Next, the user initiates the ML pipeline depicted in \autoref{fig:system}\clc{B}. 
This generates saliency measures for each of the three modalities (i.e., feature, region, and subject) through the process described in \autoref{sec:MLpipeline}.
The performance of the ML model used in this process is summarized in \autoref{fig:system}\clc{B}.
Afterward, the average and standard deviation of the saliency measures for each modality are displayed in \autoref{fig:system}\clc{C}. 
Based on the interactive exploration performed with these modules, the system updates visualizations of the associated information and brain fiber tracts (\autoref{fig:system}\clc{D}, \clc{E}). 
By reviewing these visualizations, the user may obtain insights into the neurodegenerative disease.

\subsection{Data Description and \textcolor{blue}{Processing}}
\label{sec:dataDescriptionProcessing}

\textcolor{blue}{Throughout the paper, we analyze the PPMI database's~\cite{marek2011parkinson} MRI scans/images (DTI and T1-weighted) of PD and HC subjects.
	However, our methods, including data processing, are generic enough to apply to other datasets.
	We first provide the details of the MRI images obtained by PPMI and then describe how we process the images for our analysis.}

\subsubsection{\textcolor{blue}{Data Description} }
\label{sec:dataDescription}
\textcolor{blue}{
	MRI parameters, such as gradient direction, $b$-value, and voxel resolution, have a crucial impact on the scalar measurements used for a clinical study. 
	To prevent errors, the MRI images provided by PPMI are collected based on standardized and strict acquisition protocols developed by the steering committee on 3T Siemens scanners.}

\textcolor{blue}{
	Each visit includes DTI and T1-weighted images. 
	For each DTI image, a 2D echo-planar DTI sequence is acquired with the following parameters: TR${=}\SI{900}{ms}$, TE${=}\SI{88}{ms}$, image matrix${=}116{\times}116{\times}72$ and voxel resolution${=}1.98{\times}1.98{\times}2\SI{}{mm^3}$, 64 gradient volumes ($b{=}\SI{1,000}{s/mm^2}$), and one non-gradient volume ($b{=}\SI{0}{s/mm^2}$). 
	The acquisition parameters for T1-weighted images are as follows: TR${=}\SI{2,300}{ms}$, TE${=}\SI{2.98}{ms}$, image matrix${=}160{\times}240{\times}256$, and voxel~resolution${=}1{\times}1{\times}1\SI{}{mm^3}$.
}

\subsubsection{Data Processing}
\label{sec:dataProcessing}
Here describe the details of data processing, which consists of three steps: fiber tracking, feature extraction, and cohort formulation. 
These are performed outside of the VA system.

\vspace{3pt}
\noindent\textbf{1) Fiber Tracking.}
This step generates white matter fiber tracts from the RAW images (from DTI to white matter fiber tracts in \autoref{fig:Extraction}). 
We first convert the RAW images from the Digital Image and Communications in Medicine (DICOM) format to the Neuroimaging Informatics Technology Initiative (NIFTI) format. \textcolor{blue}{Then, we perform MRI data denoising and preprocessing, including eddy-current induced distortion correction, motion correction, and susceptibility induced distortion correction, using ``dwidenoise'' and ``dwipreproc'' scripts in MRtrix3,  which is a recommended data cleaning process that uses FSL's ``eddy''~\cite{andersson2016integrated},``toppup''~\cite{andersson2003correct}, and ``applytopup''~\cite{smith2004advances} tools. This can reduce artifacts in MRI images and address many additional effects of noise during brain fiber reconstruction, such as the bias of fiber orientation estimation and error tracking of bifurcated fibers.}
\sout{Then, we fix magnet inhomogeneity (e.g., intensity loss and blurring) and perform image correction (e.g., eddy current correction and head motion correction) using the standard ``recon-all'' script in FreeSurfer~\cite{freesurfer}.} Afterward, we align the T1-weighted images to the DTI images (intra-subject registration) using FSL~\cite{fsl}. 
\textcolor{blue}{
	Intra-subject registration reduces the distortion in the anatomical structure of fibers extracted from the region of interest (ROI) in a subject.
	We also perform inter-subject registration using FSL, which applies the standard template (MNI152) to each of the subject’s MRI images.  After intra-registration and inter-registration, we perform brain parcellation using FSL, which splits the brain into regions.
	The parcellation is based on FreeSurfer’s default atlas (the Desikan/Killiany cortical atlas), which consists of 42 cortical regions\cite{desikan2006automated}.}
We then perform brain fiber tractography using a state-of-the-art framework~\cite{SMITH20121924}, which can facilitate biologically plausible fiber reconstruction and provide anatomically reliable brain fiber tracts.
\textcolor{blue}{Afterward, by referring to the brain parcellation information, we can categorize fiber tracts and obtain the corresponding features at the whole-brain level and brain region level.}

\vspace{3pt}
\noindent\textbf{2) Feature Extraction.}
We extract fiber features from the constructed brain fiber tracts (\autoref{fig:Extraction}) and diffusion tensors, using MRtrix3~\cite{TOURNIER2019116137} and FreeSurfer~\cite{freesurfer}.
The features fall into two categories: tract-based and tensor-based. 
The former measures regional fiber structures (e.g., density and length) while the latter measures water diffusivity patterns based on a tensor model. Evidence suggests that both categories are affected by neurodegenerative disease~\cite{sundaram2008diffusion,ji2015white}. 

The diffusion measures we compute with FreeSurfer include: the raw T2 signal (S0), the eigenvalues ($\lambda_1, \lambda_2$, and $\lambda_3$) representing diffusion in the directions of each of the three eigenvectors of the diffusion tensor, fractional anisotropy (FA), mode of anisotropy (MO), and mean diffusivity (MD). With MRtrix3, we also compute the other metrics, including radial diffusivity (RD), relative anisotropy (RA), axial diffusivity (AD), and the Westin metrics (linearity (CL), planarity (CP), and sphericity (CS)). O'Donnell et al. provide the definitions of these measures~\cite{o2011introduction}.

\sout{Fibers are bundled based on neuroanatomical labels assigned using a 42-region brain atlas.}
We use MRtrix3 and FreeSurfer to bundle fibers based on which  cortical regions are passed by each fiber.
Here, we use FreeSurfer to apply the cortical structure parcellation, which assigns a neuroanatomical label to each cortical region. 

Also, since PD has been reported to start from one region and then spread to others, we further divide the bundles into two categories: intra- and inter-connects by referring to the information of passed  cortical regions\sout{(connecting to same and separate regions, respectively)}.
Intra-connects (or intra-parcel connections) represent connections that both start and end within the same  cortical region while the inter-connects (or inter-parcel connections) represent the connections that start and end different  cortical regions.
Each  cortical region has both intra- and inter-connects.
Note that this definition of inter-connects does not refer to fibers that connect the two hemispheres (e.g., commissural fibers).

Also, since cortical asymmetry and hemispheric predominance have been discovered in neurodegenerative disease~\cite{scherfler2012left}, the features `Delta-LR' are extracted to represent asymmetry between the left and right hemispheres. The tract-based features include two categories:  cortical region measures and whole brain measures. The former includes the number of fibers, average fiber length, intra- and inter-fiber numbers, intra- and inter-fiber lengths, and `Delta-LR' average fiber length. The latter includes \textcolor{blue}{the number of association fibers}, projection fibers, commissural fibers, and the fibers in each brain lobe. 
Tensor-based features are averaged over the different bundles of fibers.

Our choice of features is motivated by literature review and to favor interpretability, follow standard conventions, and support fiber-tract-based analysis (\textbf{DG1}).

\vspace{3pt}
\noindent\textbf{3) Cohort Formation.} 
This step formulates cohort data (the table in \autoref{fig:Extraction}) from all scans and their attributes, including the extracted features and the corresponding demographics (age and gender) as well as their annotations, including a label of their brain status (e.g., PD or HC) and visits of scanning MRI. 
Each subject has multiple scans if they have multiple visits.
The formulated cohort data is used in the ML learning pipeline described below.
Note that we only use the label and extracted features to train the ML models, while the demographics and visit dates provide context when displaying the ML results.

\subsection{Machine Learning Pipeline}
\label{sec:MLpipeline}

The goal of our ML pipeline is to guide the user to effectively explore the data by providing measures of saliency for each feature, region, and subject (\textbf{DG1}).
The feature saliency indicates how strongly the corresponding aspect (e.g., fiber length) relates to, for example, the differences of scans with different labels (i.e., PD or HC). 

\begin{figure}[t]
	\centering
	\includegraphics[width=\linewidth]{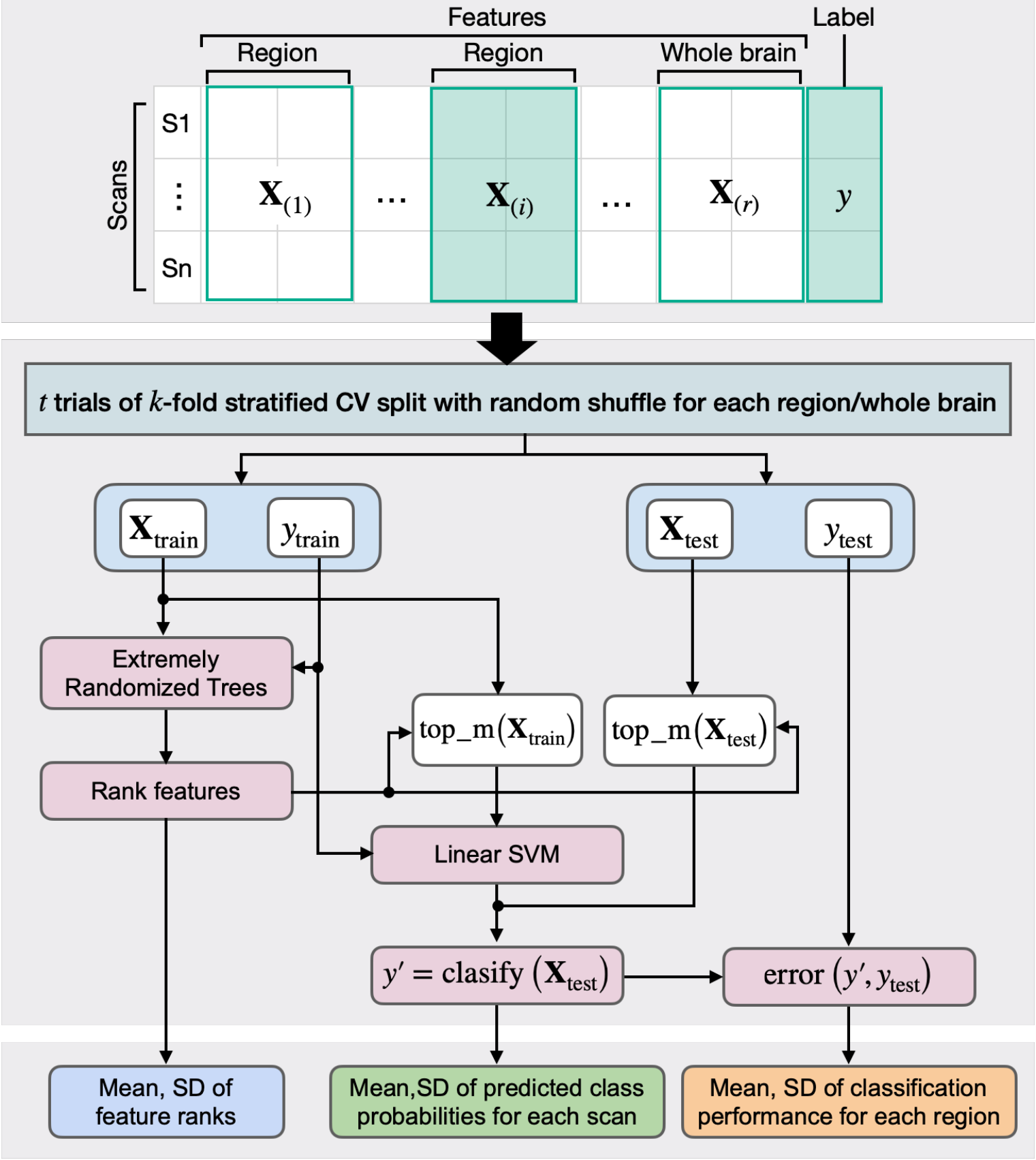}
	\caption{
		The ML pipeline. The top is the input of the ML pipeline, which is the matrix for each brain region. The middle is the core of the ML pipeline that executes binary classification in each fold of $k$-fold CV. The bottom is the output of the ML pipeline, which are feature scores, region scores, and subject/scan class probabilities.}
	\label{fig:MLpipeline}
\end{figure}

The ML pipeline, corresponding to ``Salience Guided Exploration Interface''  in \autoref{fig:system}, is described in detail in \autoref{fig:MLpipeline}. 
The pipeline's input is the cohort data generated in \autoref{sec:dataDescriptionProcessing} and the outputs are feature scores, region scores, and subject/scan class probabilities. 
The whole pipeline is executed inside CV iterations. 
In each CV iteration, we execute feature ranking and binary classification to obtain the saliency measures.
Then, we produce the averages and standard deviations of the scores over all iterations as the final outputs. 
In the following, we describe the details.

\subsubsection{Cross-Validation (CV)}
\label{sec:CrossValidation}

Over-fitting a model to data samples can reduce its generalizability to an unseen population. 
The severity increases with data complexity and limited size. 
For example, more features create a greater risk that by-chance fluctuations could discriminate target labels~\cite{krawczuk2016feature}. 
An over-fitted model learned leads to what is called generalization error due to variance.
On the other hand, an under-fitted model reduces the error due to variance but increases what is called generalization error due to bias.
The conflict between these two error terms is called the bias-variance trade-off. 
This issue is an important factor for designing our ML pipeline.

Estimating generalization error (i.e., validation) is done by resampling data into separate test and train samples. Bootstrapping (repeated sampling/testing with replacement) is a good variance estimator; however, it tends to poorly estimate bias. 
Another approach is CV, which samples without replacement. $k$-fold CV is a popular variant, which splits data uniformly into $\nFolds$ subsets and rotates their role as a test set $(\nFolds - 1)$ times. 
CV is good to estimate bias but tends to be sensitive to variance due to a dependency on the partition. 
Stratification (optimizing representativeness between subsets) helps reduce this sensitivity.
Another CV variant, repeated randomized test-train split, improves variance estimation. 
For neuroimage data, this variant has shown to work better than $k$-fold CV ~\cite{varoquaux2017assessing}. 
This makes sense since variance is a major problem in neuroimage data due to high complexity and small sample sizes. 

One of our objectives is subject-level exploration (\textbf{DG1}) using probabilistic predictions. 
However, both bootstrapping and repeated randomized CV cannot guarantee that each scan appears in a test set an equal number of times. 
Standard $k$-fold CV guarantees this but may suffer from sensitivity to variance (which is a particular problem in our domain). For these reasons, we use an extension of $k$-fold CV, which is performed $\nTrials$ times with randomization, resulting $\nTrials \times \nFolds$ iterations in total. 
This allows equal testing of scans ($\nTrials \times (\nFolds -1)$ each), and also supports good modeling of the error due to variance. 

Choice of $\nFolds$ determines the relative sizes of test and train sets at each CV iteration; a smaller $\nFolds$ results in a lower representation of the variance in a test set but increases the representativeness of the variance in a train set. 
A larger $\nFolds$ also results in more overlaps in the train sets between each CV iteration. 
The larger train sets tend to give a better estimation of the overall performance, and the overlap tends to cause more consistency between iterations; however, in turn, they may underestimate the variance (giving a false sense of stability). 
When having a small number of samples, it is also important that the test sets are representative, which implies a small enough $\nFolds$ should be chosen. 
Based on our goals, the literature, and experiment, by default, we use $\nFolds = 5$, which we have found to strike a good balance. 
We also set $\nTrials=10$ as a default to add a sufficient randomization effect while avoiding a high computational cost.
$\nFolds$ and $\nTrials$ can be adjusted within the system based on the user's need.

\subsubsection{Binary Classification}
\label{sec:binary_classification}

This stage relies on a binary classification model that learns a function $f(\bm{X}) = \hat{\bm{y}}$, where $\bm{X}$ is a matrix of the cohort data (\autoref{fig:Extraction}), with which rows and columns represent scans and attributes respectively, and $\hat{\bm{y}}$ is a prediction as to the true class labels, $\bm{y}$, that the scans belong to. 
In our case, we obtain probabilistic predictions as to whether the scan belongs to the disease (PD) or healthy (HC) group, which are used as saliency measures representing an estimation of how closely the scan exhibits patterns that are associated with the disease in the given features. 
The probabilities are then thresholded at $0.5$ to obtain the binary prediction.

Since the input $\bm{X}$ is a standard form that is compatible with many classification models, we can use many different models (e.g., SVM, decision trees, and neural networks). 
As a default model and the one used through this paper, we use a linear SVM. 
This model is popular due to its high performance with various data (including neurological data~\cite{zhang2015detection,dinov2016predictive}), robustness against overfitting, and ability to return class probabilities (rather than just predictions). 
With all of these qualities, a linear SVM is a good model for our domain and the objectives stated in \textbf{DG1}. 
One could also train neural networks to take the images directly as inputs rather than the extracted features. 
However, it is not clear if such an approach would be an improvement and the results become difficult to interpret since the features tend to be more abstract.
As our goal is to support further research through exploratory analysis and hypothesis generation rather than just prediction, we should be able to understand the results and connect them to domain knowledge.
This importance of interpretability is also stressed by feedback from experts (e.g., E4's feedback in \autoref{sec:ExpertFeedback}). 

During the CV iterations, the model is trained on a train set $\bm{X}_\mathrm{train}$, and tested on a test set $\bm{X}_\mathrm{test}$. 
Instead of using all the extracted features to train and test the model, we use the top-$\nTopFeats$ selected when performing the feature exploration to avoid overfitting (see ``Feature Exploration'' described below).
The performance of the model is measured by comparing the class predictions $\hat{\mathbf{y}}$ with the true labels $\mathbf{y}$.

\vspace{3pt}
\noindent\textbf{1) Feature Exploration.}
As described in \autoref{sec:CrossValidation}, learning from small-sample, high-dimensional data incurs a risk of false insight. The more features you have, the larger the chance that irrelevant noise patterns in the features correlate with the targets in a training set. 
One way to address this issue is through feature reduction. 
A rule of thumb is to use up to between $\sqrt{\nScans}$ and $\nScans$ features (note: $\nScans$ is the number of samples/scans), depending on how correlated they are~\cite{hua2004optimal}. 
Dimensionality reduction is one way to reduce the feature space.
However, we use the given features as they are and discard less useful ones (i.e., feature selection) when training the model as these features are easier to interpret than ones produced by dimensionality reduction. 
One common feature selection method is to use statistical significance testing between each feature and the target class. These measures are familiar to scientists and relatively easy to interpret. 
The downsides are that multivariate correlations are ignored, and the feature ranking can fluctuate widely between CV iterations. Another popular method is recursive feature elimination, which eliminates one feature at a time if the inclusion of the feature in the prediction does not improve the performance. 
However, this approach may keep too many features when they slightly improve the performance of the model and the run time can be extreme.
More importantly, it does not provide saliency measures. 
Our objectives require saliency measures---ideally stable ones. 
That is, the saliencies should not depend too much on fluctuations in the data.

Ensemble-based methods fulfill these requirements~\cite{saeys2008robust}. 
We specifically use Extremely Randomized Trees~\cite{geurts2006extremely}, a variant of random forests with added levels of randomization to reduce sensitivity to variance. 
As in \autoref{sec:CrossValidation}, the features are scored within the CV iterations. 
Finally, the averages and standard deviations over all CV iterations are used for prioritized VA and uncertainty awareness. 
Based on this average score, we rank the features. 
We also utilize the ranks to select the top-$\nTopFeats$ features that are used to train and test the linear SVM.  
Also, we report $p$-values from the Mann-Whitney $U$ test (a non-parametric test of statistical significance).
The feature scores and their uncertainties populate the feature exploration module as a table view (\autoref{fig:system}\clc{C}).
The user can then investigate individual features with the other linked views by selecting them from this view. 

\begin{figure}[t]
	\centering
	\includegraphics[width=\linewidth]{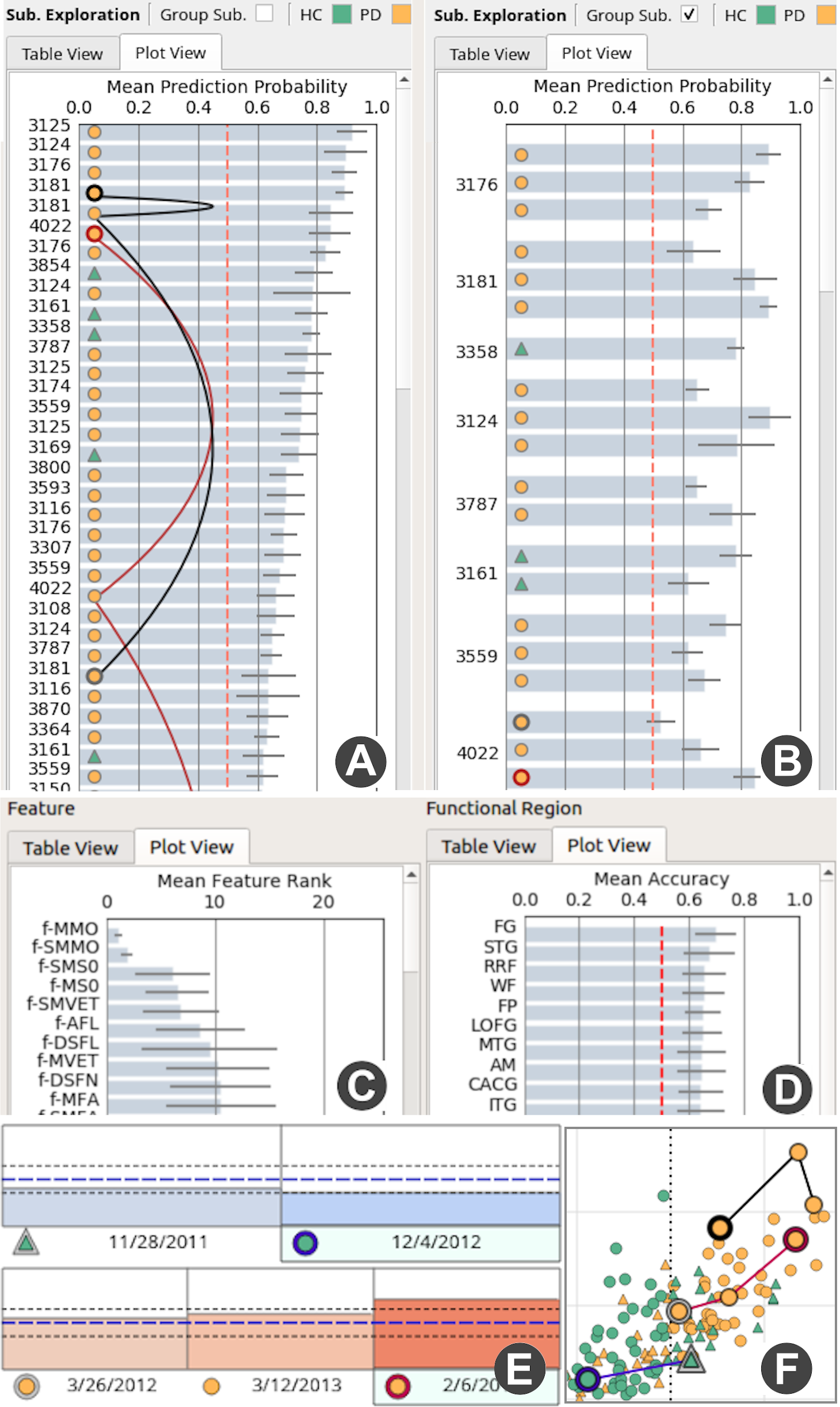}
	\caption{
		The plot views of the exploration modules and across-view encodings.
		(\glf{A}), (\glf{C}), (\glf{D}) are the plot views for the subject, feature, region exploration modules, respectively. 
		In (\glf{B}), by clicking the checkbox, `Group Sub.', the order of rows is updated to group the same subject. 
		In the 3D fiber rendering (\glf{E}) and linked visualization (\glf{F}) modules, the same glyph-based encoding with the subject exploration module is used.
	}
	\label{fig:subjectsUncertainty}
\end{figure}

\vspace{3pt}
\noindent\textbf{2) Brain Region Exploration.} After the stages above, the averages and standard deviations of classification performance have been obtained for each of the feature sets derived from the different brain regions. 
These measures are displayed in the table view to guide exploration (\autoref{fig:system}\clc{C}, \autoref{fig:subjectsUncertainty}\glf{D}).

Since each region has its own set of features and feature saliency measures, when a region is selected, all other related views are updated (e.g., the feature exploration module). 
One design consideration is whether to compute the saliencies for all regions at once or on demand. 
To make the VA process interactive and support easy non-linear exploration (\textbf{DG4}), we choose to do it all at once as this way can avoid waiting time when the user interactively explores different brain regions. 
The whole computation is done in parallel at a process level, with each process evaluating a separate region. 
Runtimes are reported in \autoref{sec:performance}.

\vspace{3pt}
\noindent\textbf{3) Across-Subject Exploration.} The average predicted class probabilities and their standard deviations over all CV iterations are also shown in the table view and guide across-subject exploration. 
This can aid the comparison and hypothesis generation (\textbf{DG3}) in a number of ways, including: model failure (why some scans do not fit the model, possible confounding conditions), model success (an obvious case of neurodegenerative expression might be found), model ambiguity (subtle expression might be found through VA or different features might be needed to disambiguate these subjects). 
In addition, we want to explore the same subject's expression across time. 
Changes in prediction across time can guide clinical analysis and disease progression.

\vspace{3pt}
\noindent\textbf{Informing Prediction Uncertainty.}
Informing uncertainty of the prediction is important when performing VA with ML to avoid gaining false insight. 
In the overall model performance view (\autoref{fig:system} \clc{B}), we display a receiver operating characteristics (ROC) curve. 
The ROC curve is used to assess the effect of the prediction threshold on performance characteristics. 
We also display the standard deviation of this curve over all CV iterations, which is useful for assessing the overall sensitivity of the model to train sets. 
After assessing these measures, the user can collapse each view to expand the screen space of the views for the exploration modules. 
Also, as mentioned, we display the standard deviation of each of the saliency measures in the exploration modules.

\vspace{3pt}
\noindent\textbf{Plot Views.}
Each of the table views can alternately be visualized as charts (\autoref{fig:subjectsUncertainty}) to graphically convey the information. 
The table and plot views can be toggled by clicking the corresponding tab at the top of each module. 
As shown in \autoref{fig:subjectsUncertainty}, we use a conventional bar chart with an error bar representing a standard deviation and sort bars based on their length (i.e., feature: rank, region: mean accuracy, subject: mean prediction probability).
In the subject exploration module (\autoref{fig:subjectsUncertainty}\glf{A}), we further show a glyph that color-codes a subject's label (orange: PD, green: HC) and encodes a correct/incorrect prediction using a different shape (circle: correct, triangle: incorrect).
Additionally, the glyph's outline color encodes the currently selected subjects (blue, red: subjects visualized in the left and right fiber views, respectively; black: mouse-hovered subject in any view).
To encode multiple scans/visits corresponding to the same subject,  we connect the corresponding rows with lines in the order of their visits for the selected subjects.
When a subject has multiple visits, to encode the order, while we apply the same outline color scheme stated above (blue, red, or black) to the last visit, we use a gray outline for the first visit and no outline color for the other visits. 
As shown in \autoref{fig:subjectsUncertainty}\glf{E} and \glf{F}, the same glyph is used for the 3D fiber rendering and the linked information visualization modules.
Also, in the subject exploration module, the user can group rows based on subject ID (\autoref{fig:subjectsUncertainty}\glf{B}).

\subsection{Visualizations for Comparative Analysis}
\label{sec:visualization}

The comparative visual analysis (\textbf{DG3}, \autoref{fig:system}\clc{D}, \clc{E}) is performed with fiber rendering and linked information visualization modules.
These views show the information related to features, regions, and scans/subjects of interest that are selected from the exploration modules.

\begin{figure}[tb]
	\centering
	\includegraphics[width=\linewidth]{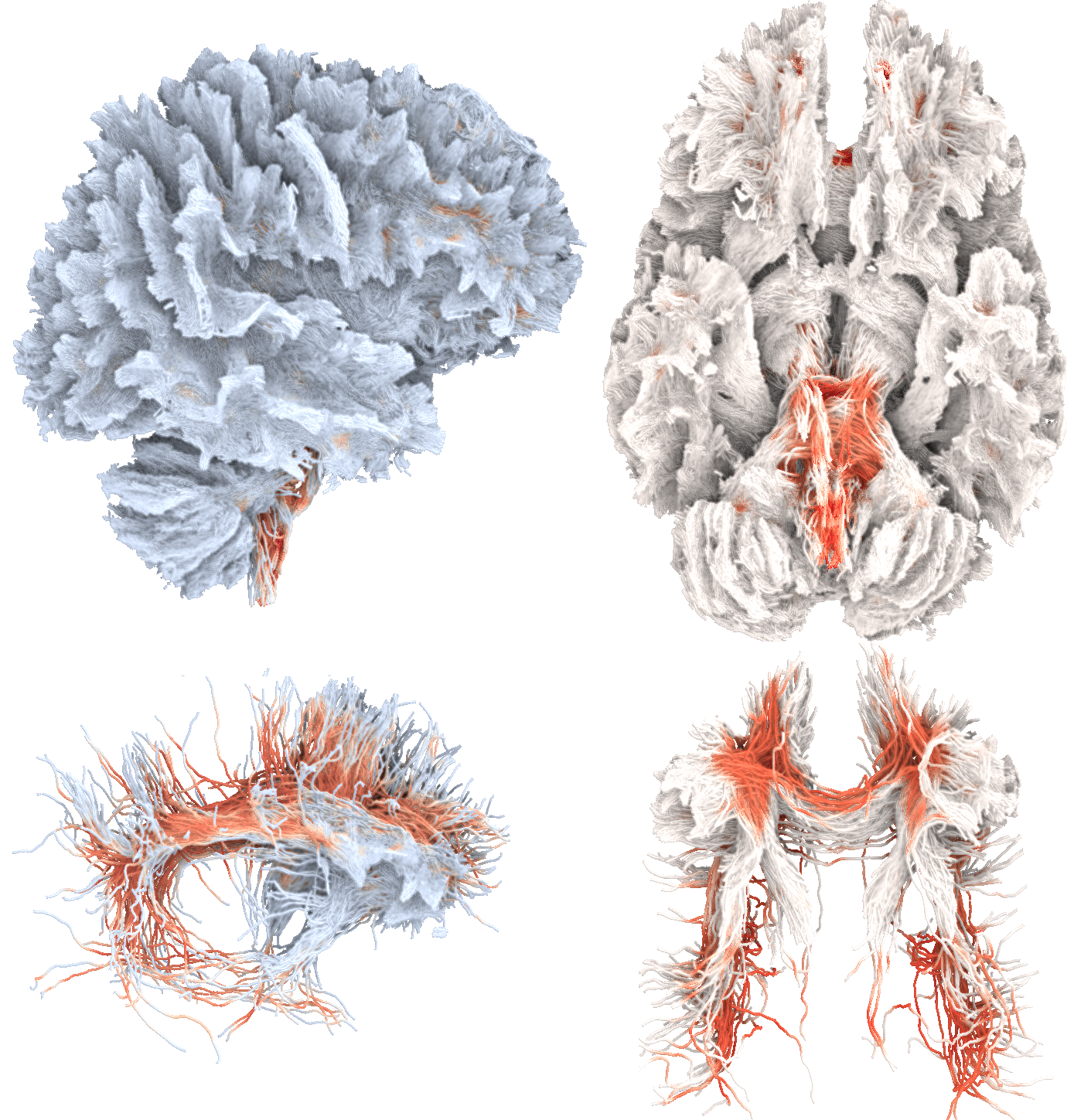}
	\caption{
		Each column shows the whole brain and one bundle at the same orientation. 
		SSAO rendering achieves a shadow effect that enhances spatial perception. 
		(Left) diverging colors encode difference from the mean of the HC group. 
		(Right) direct color mapping.}
	\label{fig:rendering}
\end{figure}

\subsubsection{Fiber Rendering}
\label{sec:rendering}

The brain fibers are rendered (\autoref{fig:rendering}) as path tubes with SSAO, which produces a high-quality visualization with an enhanced spatial perception~\cite{mittring2007finding, eichelbaum2013lineao}. 
The path tubes are constructed on the fly through the GPU rendering pipeline in the geometry shader. 
This allows the path tubes to be constructed and rendered quickly with an interactively adjustable radius without additional memory overhead. 
The scans/subjects, regions, and features selected from their respective exploration modules, automatically determine which fibers are rendered and which features are used for color mapping. 
For example, when one brain region is selected, this view only renders the fibers related to the selected region.
In addition to direct color mapping without value scaling, we provide two scaling options: contrastive color mapping---using the difference from the mean value of all the given measures in a specific \textcolor{blue}{brain} region over the entire HC group---to emphasize anomaly and logarithmic scaling to better reflect subtle value differences. However, it is important to understand that one cannot find a direct fiber-to-fiber correspondence between subjects. 
These design decisions reflect \textbf{DG2} by providing a high-quality visualization of fiber-microstructure with interactive frame-rates for large sets of brain fibers. 

The 3D fiber rendering module facilitates comparative analysis together with the linked information visualization module. 
To aid comparison, we provide two views (as shown in \autoref{fig:system}\clc{E}) and use the same colormap across all the selected scans from the other modules by referring to a global value range across the scans.
Also, the contrastive color mapping helps supports the comparison of the individuals against the HC group, which may better emphasize important differences. 
The two views' viewports can be synchronized to pan, rotate, and zoom together. 
The module also shows the subject information, such as ID, age, gender, and label, as shown in \autoref{fig:system}\clc{E}.
In addition, the bar chart placed at the top-right of each view (\autoref{fig:system}\clc{E}, \autoref{fig:subjectsUncertainty}\glf{E}) shows the mean value of the selected feature for each scan/visit. 
The bar chart also shows two different dotted lines. 
The blue line represents the average of the HC group while the black line shows the standard deviation.

\subsubsection{Linked Information Visualization}
\label{sec:comparison}

We include a set of linked information visualizations to support various comparisons. 
This includes the scatterplot, parallel coordinates, dimensionality reduction, histogram, trend, covariance matrix, and correlation matrix views. 
These views depict the relationships/distributions among selected attributes (including the extracted features, demographics, predicted class probabilities) and/or scans. 
For example, the parallel coordinates view informs multiple selected attributes' values for each scan while the dimensionality reduction view depicts each scan's similarity computed based on the attributes' values as the spatial proximity.
The implemented views provide commonly used visualizations (e.g., a heatmap showing a covariance matrix).
We demonstrate the usage of these views in the case studies in \autoref{sec:cases}. 

While we use a standard visualization for each view, we design a method to decide the order of features shown in the parallel coordinates, covariance matrix, and correlation matrix as the order is important to help the user visually find patterns~\cite{1382895}.
We apply hierarchical clustering to group correlated features using the Louvain community detection algorithm~\cite{donetti2004detecting}. 
The same sorted order is applied to all the three views to assist the mental map~\cite{1215004}. 

\section{\textcolor{blue}{Evaluation}}
\label{sec:cases}

\textcolor{blue}{To evaluate our system, we provide case studies, expert feedback, and performance evaluation.As with \autoref{sec:methods}, we use the PPMI database~\cite{marek2011parkinson} for our evaluation.} Five experts (\Expert{1}--\Expert{5}) with different professional backgrounds are involved in the evaluation of our system. 
\Expert{1} is a radiologist who focuses on 3D visualization of medical images and brain disease analysis of high-resolution brain fiber tracts. 
\Expert{2} is a neurologist who specializes in treating motor disorders, such as stroke, PD, and epilepsy. 
\Expert{3} is a physician of neurology, mainly engaged in cerebral stroke and Alzheimer's disease clinical/research work. 
\Expert{4} is a neuroscientist with expertise in statistical analysis, tractography, and neurodegenerative disease. 
\Expert{5} is a neurologist at a children’s hospital specializing in diagnosis and treatment of congenital nervous system malformations, and voxel-based morphometry analysis of Huntington’s disease. 
\Expert{1}, \Expert{2}, and \Expert{3} were involved in the case studies while \Expert{4} and \Expert{5} only provided qualitative feedback. 
One challenge in evaluating our system is that none of the consulted domain scientists is an expert in all the three areas (brain fibers, PD, and ML).
We should also point out that we are not drawing scientific conclusions about PD in our case studies. Instead, we focus on demonstrating the usage of our system and the types of insights it can provide.

\subsubsection{Background Knowledge of PD}

\noindent PD is a neurodegenerative disorder characterized by loss of dopamine neurons in the substantia nigra (SN) brain region~\cite{prakash2012asymmetrical}. 
The dopaminergic function of the nigrostriatal pathway reduces with the depletion of these neurons as well as the neural fibers that link the SN to other subcortical regions, such as putamen and caudate \cite{zhang2015diffusion}. 
Each region in the midbrain is affected based on the anatomical correlation with SN and the severity of the region reflected by the \textcolor{blue}{anatomical}  relationship with SN.
It is recognized to be a brain-wide neurodegenerative process that spreads up from the brainstem into the cortex, as originally suggested by Braak and Braak~\cite{yau2018network}. 
PD also increases rapidly with age---while low incidence before $50$, most cases are found around $80$~\cite{ASCHERIO20161257}. 
Once the fibers are detected by tractography, it is useful to assess the diagnosis with the brain fiber pathways. 

\begin{figure*}[t]
	\centering
	\includegraphics[width=\linewidth]{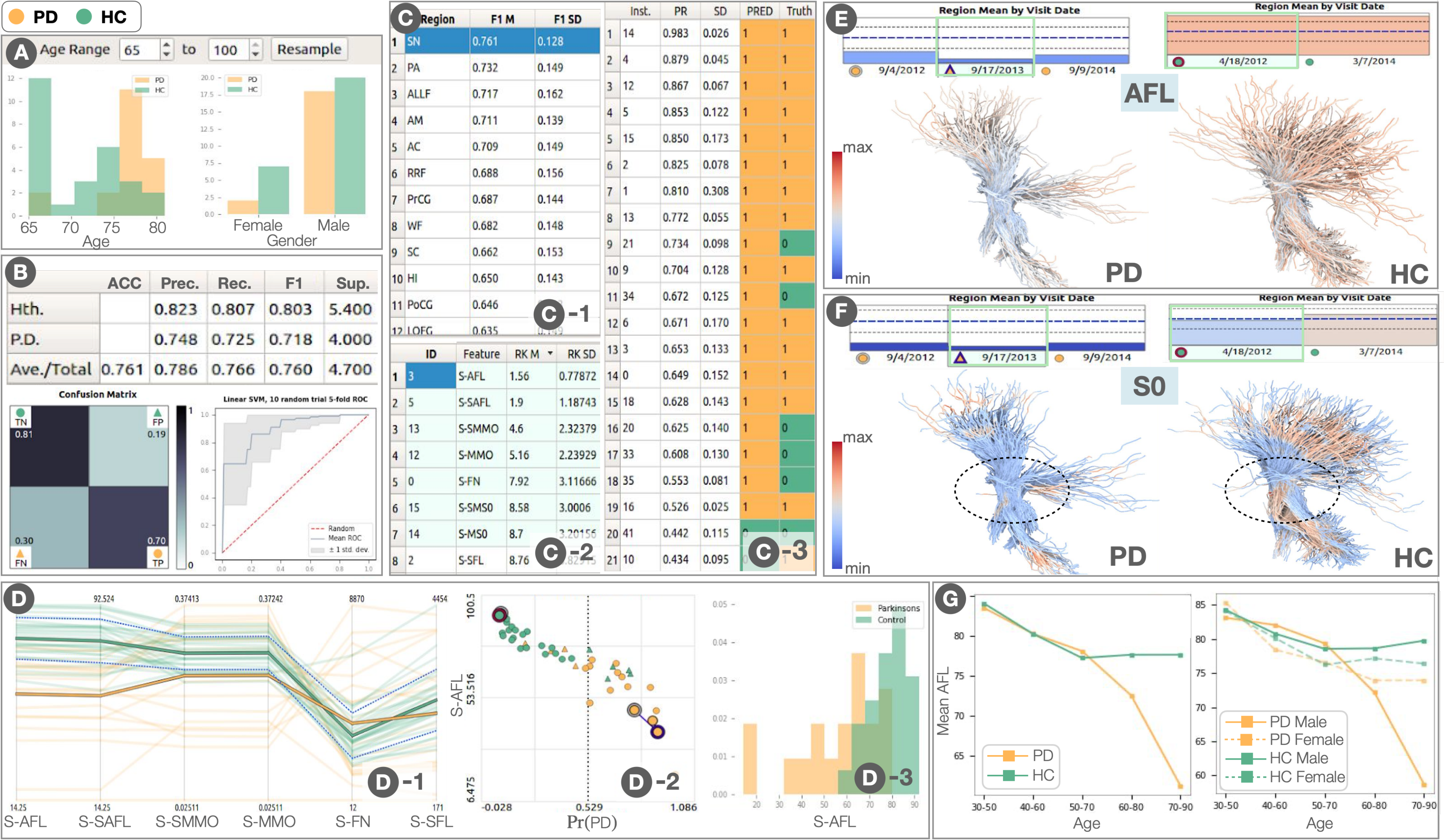}
	\caption{Performing Case Study 1 with the VA system. 
		Here we show the cohort selection (\glf{A}), ML (\glf{B}), exploration (\glf{C}) (1: region, 2: feature, 3: subject), linked information visualization (\glf{D})(\glf{G}) , and 3D fiber rendering  modules (\glf{E})(\glf{F}). 
		The information visualization module is used to show the parallel coordinate (\glf{D}-1), scatterplot (\glf{D}-2), distribution (\glf{D}-3), trend (\glf{G}) views.
		(\glf{E}) and (\glf{F}) shows the fiber tracts colored by AFL and S0 values, respectively.
	}
	\label{fig:SNSteps}
\end{figure*}

\subsubsection{Study 1: Exploration of the SN Region}

Based on the background knowledge of PD, we decide to explore fiber bundles with streamlines starting from the SN region; our focus is on trends with age and on spatial feature patterns from the elderly subgroups (where significant symptoms have been reported). 
First, from the entire scans we have, we investigate the relationships between age and the average fiber length (AFL) in the SN region by using the trend view in the linked visualization (\autoref{fig:SNSteps}\glf{G}). 
From the right plot where we show the association between AFL and age for each group and gender, we see a clear drop in AFL in the elderly male subjects with PD. 

From the trend above, we decide to review the elderly subjects in more detail. 
To do so, we resample the dataset into only elderly subjects (i.e., age 65--100), as shown in \autoref{fig:SNSteps}\glf{A}. 
One issue is that we have limited scans in this age group, including a significant lack of female subjects, as shown in the distributions in \autoref{fig:SNSteps}\glf{A}. 
This helps notify the limitations of the data and uncertainty implications before proceeding in the ensuing exploration. 
For this analysis, we decide not to use resampling that generates the balanced age and gender distributions (which is supported by the system) because this can produce a very small resampled data when using with the restriction on age, and disallows the detailed exploration of the discarded subjects. 
The ML pipeline executes the linear SVM with the resampled data, resulting in 76.1\% average accuracy and 0.760 average $F_1$ for the SN region over all CV trials  (\autoref{fig:SNSteps}\glf{B}). 

Afterward, we review the ranks of SN features with \autoref{fig:SNSteps}\glf{C}-2. 
The high-rank features include the average fiber length (AFL), the mode of the anisotropy (MO), the fiber number (FN), and the raw T2 signal with no diffusion weighting (S0), each for the entire SN (i.e., both inner- and inter-connects) and the inner-connect fibers (e.g., AFL: entire, SAFL (or self-connect AFL): inner-connect). 
With the parallel coordinates view (\autoref{fig:SNSteps}\glf{D}-1), we show a general value trend of the top-7 features. 
We observe that several features' PD group average (thick orange line) are outside of the range of the HC group average plus/minus its standard deviation (the range between the blue dotted lines around the thick green line).
As AFL shows the most clear difference between the PD and HC group average, we compare its value distributions with the histograms (\autoref{fig:SNSteps}\glf{D}-3), where the HC group tends to have higher values than the PD group.
Also, we show the scatterplot of AFL and the prediction probability of PD ($\mathrm{Pr}(\mathrm{PD})$) to know the relationship between AFL and the trained classifier (\autoref{fig:SNSteps}\glf{D}-2). 
A linear-decreasing trend seen in the plot clearly suggests that the trained model exploits a correlation between PD and AFL. 

To review the differences in the fiber tracts between PD and HC subjects, from the subject exploration module (\autoref{fig:SNSteps}\glf{C}-3), we select each group's representative subject who has the high prediction probability and are predicted correctly. 
We first compare their AFL values with the contrastive color mapping, as shown in \autoref{fig:SNSteps}\glf{E}. 
The HC subject's nigrostriatal fibers are mostly light red, indicating higher than the average length, while the PD subject has more blue fibers, indicating lower than the average length. 
We also compare S0 values (\autoref{fig:SNSteps}\glf{F}), which is a feature of DTI and frequently used for white matter analysis\cite{zhang2014language,inano2014voxel}.
Here, the PD subject's S0 values around the center of SN (annotated with the dotted-black ellipses) are higher than those of the HC subject. 
Also, the information of the subjects' multiple visits (at the top of each view) shows these subjects have fairly consistent values across their visits.

\Expert{1}--\Expert{3} provided positive opinions on the functionalities used to explore the patterns of SN fibers and expressed the intuitiveness of the 3D fiber rendering module to understand the feature value differences. 
They pointed out that the phenomenon of nerve cell death in the SN region with the age and disease progression is reasonable. 
They also reminded us that this is a manifestation of PD but may not help explain the underlying mechanisms of the cause of PD.

We also note the limitations of this case study.
Based on the findings from existing works, we originally expected the SN regions could show significant differences between PD and HC (even in any age group). 
However, with the data we used, we were only able to find the clear differences in the elderly groups. 
As we selected a small age range, this analysis might do not include sufficient subjects to make confident inferences. 
Also, we find high variation in the fiber features among individuals, making concluding the clear associations difficult. 
These problems would be addressed in further work when having a larger collection of data.
Beyond the microstructural differences of the nigrostriatal tracts in PD subjects~\cite{zhang2015diffusion}, T2 hypointensity has been reported as a sensitive measure that is caused by the iron accumulation in the substantia nigra in PD~\cite{ollivier2018neuroimaging}. 
Low S0 values can be a sign of high iron concentrations. 
In the data we used, we found that the overall PD group has a lower mean value of S0 in the SN region except for some outliers with higher values. 
Even though our analysis result is inconclusive, the gained insights could be useful to direct further research with hypothesis generation. 
For example, \Expert{3} said that the iron accumulation in one's brain may be caused by many different factors but its associations to the pathogenesis of PD are not known yet. 
To study the relationships with PD, it is worth investigating how the iron has accumulated.

\begin{figure*}[t]
	\centering
	\includegraphics[width=\linewidth]{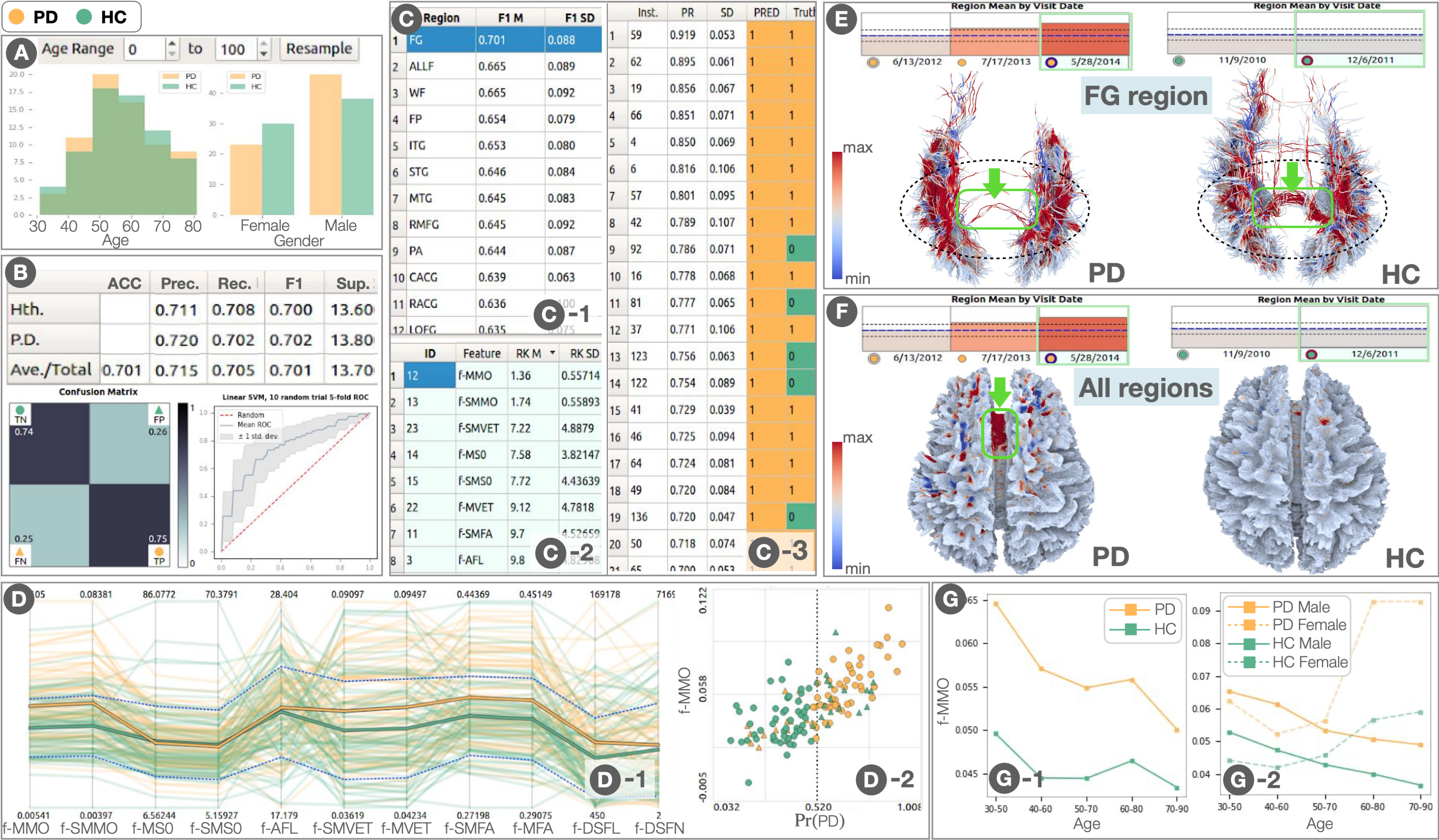}
	\caption{
		Performing Case Study 2 with the VA system. 
		All the views are the same with \autoref{fig:SNSteps} but show the information related to Case Study2.
	}
	\label{fig:fusformSteps}
\end{figure*}

\subsubsection{Study 2: Exploration Based on the ML Suggestions}
\label{sec:case2}

Unlike the first case study, we analyze the data while fully utilizing the saliency measures produced by the system and do not rely on the assumption from past knowledge (e.g., in the first study, we have focused on the SN region). 
We are interested in exploring the brain regions that our system indicates the best predictors of PD, the top-performing features in those regions, detailed patterns of those features in the physical space, and then trends with age and gender.  

We begin by applying the ML pipeline to the entire data, including $68$ PD and $68$ HC subjects (their age and gender distributions are shown in \autoref{fig:fusformSteps}\glf{A}). 
The resultant overall prediction accuracy is 70.1\% and $F_1$ is $0.701$ (\autoref{fig:fusformSteps}\glf{B}). 
The saliency measures for the three modalities are displayed in \autoref{fig:fusformSteps}\glf{C}. 
In the region exploration module (\autoref{fig:fusformSteps}\glf{C}-1), the system suggests that the fusiform region (FG) is the best predictor. 
Thus, we select FG to explore the saliency measures of features within FG. 
From the result shown in \autoref{fig:fusformSteps}\glf{C}-2, the top-salient feature within FG is the mean of the mode of anisotropy (MMO).
The mode of anisotropy (MO) is a shape metric representing diffusion patterns. 
We should note that, during an interactive exploration, we observe that MMO is the top-salient feature for many other regions and the whole brain. 
This can happen because the decrease of MO is often used as one of the indicators of PD~\cite{prange2019early}.

We move on to the group-level analysis using the linked information visualization module. 
We first use the parallel coordinates (\autoref{fig:fusformSteps}\glf{D}-1) to see the trends of the top-11 features selected from the feature exploration module. 
For all the features, we can see a high variation among scans (one scan is represented as one polyline in the parallel coordinates).
While we can see many overlaps between PD and HC for most of the features, MMO placed at the far left shows a clearer difference between them (many PD subjects have a higher MMO value). 
From the scatterplot of MMO and $Pr($PD$)$ (\glf{D}-2), we can see a positive linear trend with a relatively clear separation between PD and HC, indicating MMO is a relatively good indicator of PD.

Afterward, we select multiple subjects from the PD and HC groups to compare their difference with the 3D fiber rendering module.
As we have already found the PD group tends to have higher MMO values (as shown in \autoref{fig:fusformSteps}\glf{D}-2), here we further inspect MO values in a fiber level.
We find that some HC subjects with a low mean MO tend to have more fibers around the area annotated with the dashed-black ellipse in \autoref{fig:fusformSteps}\glf{E} (note that while we only show one example here, we see other subjects have a similar pattern).
Another common structural characteristic is that some PD subjects have fewer fibers between the occipital and parietal lobes (i.e., the area annotated with the green arrows in \autoref{fig:fusformSteps}\glf{E}). 
This is an interesting finding for further investigation because the reduced fiber density connecting between the occipital and parietal lobes has been suggested to be a possible sign of diseased brain atrophy in PD~\cite{10.1093/brain/awh088}. 

Besides focusing on FG, we also visualize the whole brain colored by MO (\autoref{fig:fusformSteps}\glf{F}). 
We notice that the PD subject shown in \autoref{fig:fusformSteps}\glf{F} (who has a high MMO) has an area filled with high values (dark red), as annotated by the green color.
Such areas where high MO values are dominant could highly influence the prediction performed by the ML model and could be also the source of higher MMO values in the PD group.
In the physiological sense, the different distributions of MO intensity indicate different patterns of water diffuse in different parts of the brain; however, imaging issues could also affect these tensor measures.

We further investigate to know whether or not those areas are the source of the higher MMO value.
Through exploring multiple subjects in the 3D fiber rendering module, we identify that the higher MMO is not caused by having slightly higher MO values across all areas but by having some specific areas with extremely higher MO values (as with the area annotated in \autoref{fig:fusformSteps}\glf{F}). 
This finding highlights that examining the tensor measure value differences by areas is important to interpret the ML results and also to know the structural differences between subjects. 
Such an analysis is greatly aided by using the 3D fiber rendering module, as we have demonstrated above.

Now, we analyze the relationships among MMO, age, and gender with the trend view, as shown in \autoref{fig:fusformSteps}\glf{G}.
From \autoref{fig:fusformSteps}\glf{G}-1, we can see the PD group has a higher MO than the HC group. 
However, as shown in \autoref{fig:fusformSteps}\glf{G}-2, the trends of the male and female subjects show different patterns: the male group has the decrease of MO by age for both PD and HC while the female group has the opposite increase trend. 
However, the limited amount of data (especially, the elderly female PD subjects, as discussed in the first case study) makes this insight highly uncertain; thus, a further investigation would be needed to confirm this finding. 

\Expert{2} showed interest in the above findings. 
Clinically, PD mainly has five symptoms: dementia, impaired balance, slowness/bradykinesia, stiffness/rigidity, and resting tremor. 
Each patient has one or more symptoms and their symptoms are different. 
The loss of connection between the occipital and parietal lobes may be related to one symptom (e.g., impaired balance) while the areas with high MO values may be associated with other symptoms. 
Also, the experts suggested co-analyzing the MO feature with the fiber loss. 
These two features can be utilized to characterize the different stages of PD as their co-joint value distribution tends to show certain different patterns between the cases where a PD patient has and does not have the impaired balance. 

\subsection{Expert Feedback}
\label{sec:ExpertFeedback}

As stated, to help evaluate our work, we sought feedback from the experts in related research fields, including statistical analysis of neurodegenerative disease,  brain tractography of MRI images, and diagnosis and treatment of PD.
Our team members conducted live interactive demos for each expert, and also shared a draft of our manuscript, and then assimilated the feedback into our design and paper.

We added several functionalities based on the interactions with the experts. 
For example, we implemented the trend view (e.g., \autoref{fig:fusformSteps}) based on \Expert{1}'s request to investigate how age and gender affect the disease.
Also, while we originally used statistical significance testing as a feature ranking method, we changed to utilizing Extremely Randomized Trees, as discussed in \autoref{sec:binary_classification}. 
Besides the advantages discussed already, part of the motivation is that significance testing can be easily misused and should be performed under careful experimental settings. 
The scores from Extremely Randomized Trees, on the other hand, can be simply interpreted as a relative score to rank features while this approach is effective as shown in the performance of the classifier.
However, \Expert{3} told that they would still like to see $p$-values. 
In their field, the $t$-test is commonly used; however, it depends on the assumption of normality, which is not guaranteed to hold for our extracted features. 
Thus, instead, we selected and incorporated the Mann-Whitney $U$-test, which is a non-parametric statistical significance test that has fewer assumptions and does not require normality. 
However, we still want to emphasize the importance of careful interpretation of the testing result. 
Some experts expressed that they were distracted by the classification performance charts, which we only use when assessing the ML performance.
Therefore, we made these charts collapsible.

\Expert{4} and \Expert{5} described their thoughts on general research challenges and directions related to our work.
\Expert{4} stated that since the fiber-based analysis of neurodegenerative diseases is still in the early stages, current research in their field is highly exploratory. 
Advanced methods such as fiber tractography are being actively studied and showing a promise for scientific discovery; however, the complexity and uncertainty they might face when using these methods  ``\textit{scares away some researchers}'' (\Expert{4}). 
Based on \Expert{4}'s experience, due to those complexities and practical difficulty in the comparison of many different individuals' fiber tracts, the research focus tends to rely more on statistical comparisons (e.g., DTI measures averaged for each region).
For \Expert{4}, those fiber-tract-based analyses are usually only initiated to investigate outliers. 
Overall, \Expert{4} stated that our VA system would enable them to frequently perform a practical qualitative analysis of the fibers.
Also, using ML to guide an exploratory VA is a new approach to them, and \Expert{4} expressed concern about a lack of ML knowledge in many neuroscientists to understand ML models and parameters.
\Expert{5} stated our VA system is innovative and useful as a research tool because it helps study neurological disease from a new perspective, such as the morphological aspects.
However, in clinical practice, \Expert{5} expressed that only well-understood and standardized markers can be translated into actionable insights.
\Expert{5} also suggested that the system can be more useful for clinicians by avoiding asking them to use the pipeline by themselves and assigning ML experts to hospitals to properly utilize the ML pipeline.
Also, as some biomarkers are critical in diagnosis and disease evaluation, \Expert{5} suggested that a promising next step could be analysis of co-occurrences of those markers and tract-based features; then, in the future, to achieve better diagnosis and evaluation, those tract-based features would be added as additional index variables in multi-modal disease severity grading systems.
We consider that these experts' comments are insightful and would be useful for VA researchers to plan their future work. 

	\subsection{Performance}
\label{sec:performance}

We evaluate the computational performance of our ML pipeline.
The most expensive computation in each CV iteration is the feature ranking using the Extremely Randomized Trees. 
The feature ranking's runtime highly depends on the number of trees/estimators to use. 
While the system uses 150 trees/estimators by default, this number can be as high as 1,500 to provide better results.
CV parameters, the numbers of folds, $\nFolds$, and trials, $\nTrials$, are also linearly correlated to runtime. 
We test the performance with multiple settings and the full set of data (136 subjects). 
We train linear SVM with the top-$\sqrt{\nScans}$ ranked features.
As an experimental platform, we use a machine running Arch Linux with a 3.60 GHz Intel Core i3-9100F and 32GB of DDR4 memory clocked to 2133 MT/s.
With the default parameters (150 estimators/trees, $\nFolds=5$, $\nTrials=10$), the runtime is 42 seconds. 
When using 1,500 estimators/trees, $\nFolds=10$, $\nTrials=10$, the runtime is 325 seconds.
These runtimes are reasonable as the execution of this pipeline is only required once (unless resampling is performed) and does not impact the interactive exploration.
	\section{Discussion and Limitations}

The ML approaches and features we use have been verified by researchers in neuroscience~\cite{mateos2018structural,tanveer2020machine,acosta2016whole,wen2016white}. 
Direct visualization of features over individual fibers can alert underlying issues in the use of the averaged features for group-level comparison (as we have demonstrated in \autoref{sec:case2}). 
This fiber-tract visualization also provides non-trivial physiological explanations that could enhance the current understanding of neurodegenerative diseases. 

As one gains insight into the important structural patterns, a rational next step would be to develop formalized descriptors that could be automatically extracted from the data. 
Currently, qualitative analysis through visualization is used to investigate the differences between individuals' brain structures, but those differences may not turn out to be statistically significant upon group-level analysis. 
One direction for further research is toward spatially invariant feature localization and extraction using deep neural networks; however, while utilizing visualization, we should address the problems incurred by complex ML models, such as explainability, ease-of-use, and standardization. 

Another limitation comes from issues with data assimilation. 
Our current system requires all brain data to be generated using identical imaging parameters and processing. 
With more reliable techniques to assimilate data from different sources, we could greatly expand the amount of data that could be used together. 
With larger data, narrowing down significant differences would be easier.
Similarly, we would be able to perform more fine-grained fiber-tract-based analysis as well as higher confident group analyses (e.g., based on age and gender).
Since the state of the art in tractography is actively evolving, in the future, there should be more standardized and well-understood methods in use, which will appease researchers who wish to compare results between studies and understand and judge the analysis results with high confidence. 
It is noteworthy that current tractography can be computationally expensive. 
With a mid-range workstation, it took us about $60$ days in total to process about $190$ brain scans using a mid-range workstation. 
Faster implementations utilizing acceleration hardware, such as GPUs, will tremendously benefit the field.
In addition, a large amount of storage space is needed to store the data. 
For example, each subject's data, including the fiber tracts and their feature measures, has the size of several gigabytes.

	\section{Conclusions}

Our system can effectively facilitate a deeper physical investigation into the statistical measures that are used by researchers to study differences between healthy control and neurodegenerative disease groups of DTI fiber tracts. 
The efficiency of the investigation is increased through intelligent guidance in the exploration process using predictive modeling, while comparative analysis is enhanced through customized interactive visualizations that are directly linked with each other and the predictive modeling pipeline. 
This set of visualizations helps provide better context to the observed differences by simultaneously expressing multiple comparative modes for analysis and emphasizing uncertainty. 
This approach can benefit neurodegenerative disease researchers by helping them easily gain a wide perspective into their data as they search for insights through an exploratory analysis process.

	
	%

	

	\ifCLASSOPTIONcompsoc
	\section*{Acknowledgments}
	\else
	\section*{Acknowledgment}
	\fi
	
	This research is sponsored in part by the U.S. National Science Foundation through grants IIS-1528203 and IIS-1741536, and the Nature Science Foundation of China through grant 61976075.
 
We thank Dr. Pauline Maillard from the Department of Neurology, the University of California at Davis, Dr. Xiufang Xu from Hangzhou Medical College, and Dr. Chao Lin from the Children's Hospital of Zhejiang University School of Medicine who provided insight and expertise that greatly assisted the research. We would also like to show our gratitude to Dr. Shunyuan Guo and Dr. Gaoping Lin from Zhejiang Provincial People’s Hospital for sharing their pearls of wisdom with us during this research.

Data used in the preparation of this article were obtained from the
Parkinson's Progression Markers Initiative (PPMI) database (\url{www.ppmiinfo.org/data}). For up-to-date information on the study, visit \url{www.ppmiinfo.org}.
PPMI---a public-private partnership---is funded by the Michael J. Fox Foundation
for Parkinson’s Research and funding partners, including AbbVie, Allergan, Avid Radiopharmaceuticals, Biogen, BioLegend, Bristol-Myers Squibb, Celgene, Covance, GE Healthcare, Genentech, GlaxoSmithKline, Golub Capital, Handl Therapeutics, Insitro, Lilly, Lundbeck, Merck, Meso Scale Discovery, Pfizer, Piramal, Prevail, Roche, Roche, Sanofi Genzyme, Servier, Takeda, TEVA, UCB, Verily, and Voyager.

	\ifCLASSOPTIONcaptionsoff
	\newpage
	\fi

	
	
	\bibliographystyle{IEEEtran}
	\bibliography{IEEEabrv,00_bibliography}
	
	%
	
	
	
	%

	
	
	
	
	\begin{IEEEbiography}[{\includegraphics[width=1in,height=1.25in,clip,keepaspectratio]{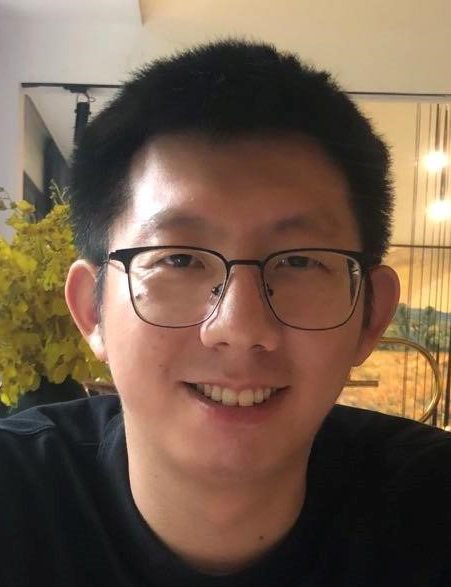}}]%
    {Chaoqing Xu} received the BS degree in automation from Zhejiang University of Science and Technology. He is a Ph.D. student in control science and engineering from the Zhejiang University of Technology, Hangzhou, China. He was a  visiting scholar at the University of California, Davis, from January 2017 to October 2018. He is currently working with Ronghua Liang. His main research interests include medical visualization and brain fiber analysis.
\end{IEEEbiography}

\begin{IEEEbiography}[{\includegraphics[width=1in,height=1.25in,clip,keepaspectratio]{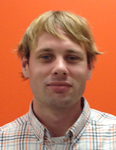}}]%
    {Tyson Neuroth} received the BS degree in computer science from the University of California, Davis. He is now a Ph.D. student at the University of California, Davis, studying computer science and visualization under Kwan-Liu Ma. His research interests include scientific visualization, high performance computing, and human-computer interaction. 
\end{IEEEbiography}

\begin{IEEEbiography}[{\includegraphics[width=1in,height=1.25in,clip,keepaspectratio]{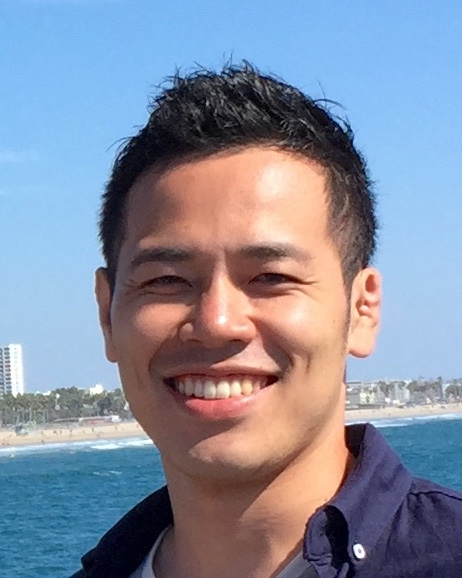}}]%
    {Takanori Fujiwara} is a Ph.D. candidate in computer science at the University of California, Davis, advised by Kwan-Liu Ma. He received his Master's degree in environmental science in 2011 from the University of Tokyo. Before UC Davis, he worked for Kajima Corporation in Japan. He works at the intersection of data science and data visualization with special interests in visual analytics of high-dimensional data and network data using machine learning. 
\end{IEEEbiography}

\begin{IEEEbiography}[{\includegraphics[width=1in,height=1.25in,clip,keepaspectratio]{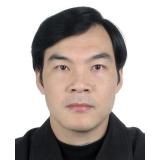}}]%
    {Ronghua Liang} received the PhD degree in computer science from Zhejiang University in 2003. He was a research fellow at the University of Bedfordshire, United Kingdom, from April 2004 to July 2005 and as a visiting scholar at the University of California, Davis, from March 2010 to March 2011. He is currently a professor of computer science and dean of the College of Computer Science and Technology, Zhejiang University of Technology, China. He has published more than 80 papers in leading international journals and conferences including IEEE Transactions on Knowledge and Data Engineering, IEEE Transactions on Visualization and Computer Graphics, IEEE Information Visualization, IJCAI, AAAI. His research interests include medical visualization, image processing, and big data visualization. He is a member of the IEEE.
\end{IEEEbiography}
 
\begin{IEEEbiography}[{\includegraphics[width=1in,height=1.25in,clip,keepaspectratio]{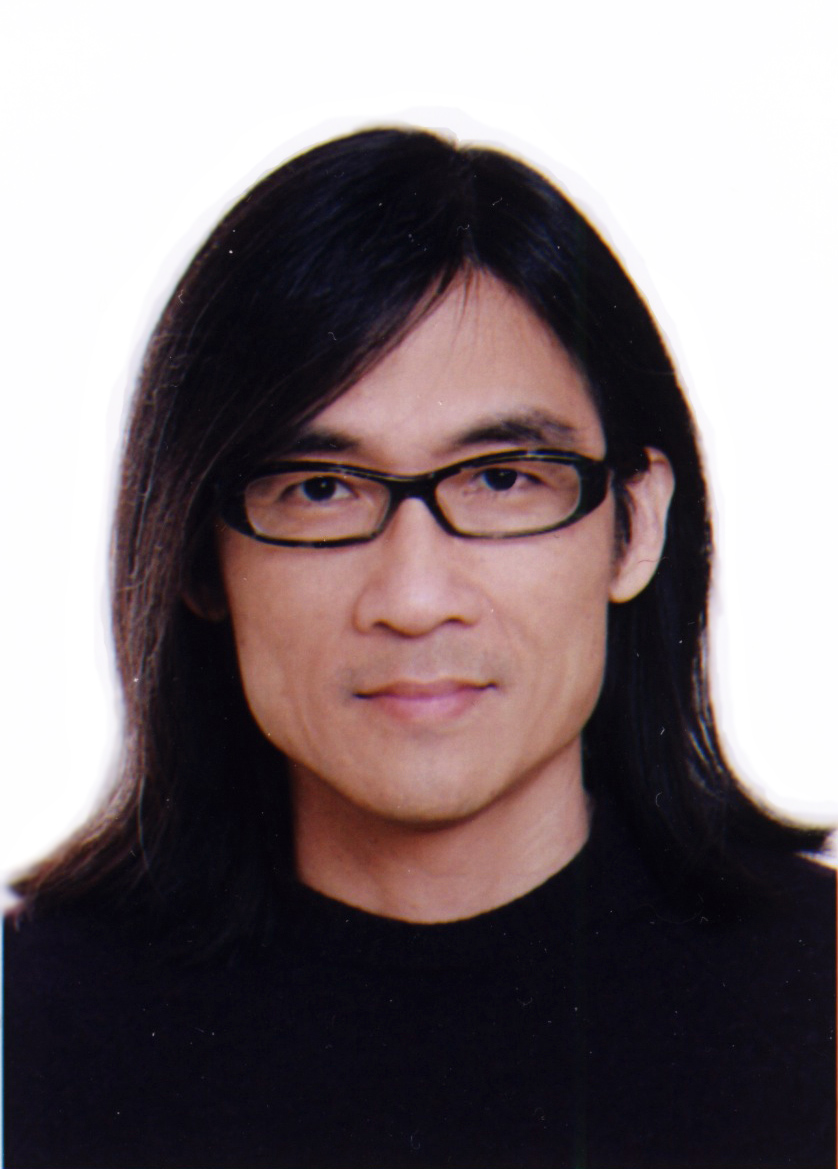}}]%
    {Kwan-Liu Ma} is a distinguished professor of computer science at the University of California, Davis. He received his PhD degree in computer science from the University of Utah in 1993, and then worked as a staff scientist at ICASE/NASA Langley Research Center before joining UC Davis in 1999.  His research is in the intersection of data visualization, computer graphics, human-computer interaction, and high performance computing.  For his significant research accomplishments, Ma received several recognitions including elected as IEEE Fellow in 2012, recipient of the IEEE VGTC Visualization Technical Achievement Award in 2013, and inducted to IEEE Visualization Academy in 2019.
\end{IEEEbiography}

	
	\vfill
	

\end{document}